\documentclass{article} 
\usepackage{iclr2015,times}
\usepackage{hyperref}
\usepackage{url}
\usepackage{amsmath,amssymb}
\usepackage{caption}
\usepackage{subcaption}
\usepackage{sidecap}
\usepackage{graphicx}

\usepackage{tikz}
\usetikzlibrary{bayesnet}

\title{Variational Recurrent Auto-Encoders}

\author{
Otto Fabius \& Joost R. van Amersfoort \\
Machine Learning Group \\
University of Amsterdam \\ \texttt{\{ottofabius,joost.van.amersfoort\}@gmail.com}
}

\iclrfinalcopy 

\sidecaptionvpos{figure}{t}
\begin{document}

\maketitle

\begin{abstract}
In this paper we propose a model that combines the strengths of RNNs and SGVB: the Variational Recurrent Auto-Encoder (VRAE). Such a model can be used for efficient, large scale unsupervised learning on time series data, mapping the time series data to a latent vector representation. The model is generative, such that data can be generated from samples of the latent space. An important contribution of this work is that the model can make use of unlabeled data in order to facilitate supervised training of RNNs by initialising the weights and network state.
\end{abstract}

\section{Introduction}
Recurrent Neural Networks (RNNs) exhibit dynamic temporal behaviour which makes them suitable for capturing time dependencies in temporal data. Recently, they have been succesfully applied to handwriting recognition \citep{graves2009novel} and music modelling \citep{boulanger2012modeling}. In another more recent development, \citet{cho2014learning} introduced a new model structure consisting of two RNN networks, an encoder and a decoder. The encoder encodes the input to an intermediate representation which forms the input for the decoder. The resulting model was able to obtain a state-of-the-art BLEU score.

We propose a new RNN model based on Variational Bayes: the Variational Recurrent Auto Encoder (VRAE). This model is similar to an auto-encoder in the sense that it learns an encoder that learns a mapping from data to a latent representation, and a decoder from latent representation to data. However, the Variational Bayesian approach maps the data to a \textit{distribution} over latent variables. 
This type of network can be efficiently trained with Stochastic Gradient Variational Bayes (SGVB), introduced last year at ICLR by \citet{kingma2013auto}, and our resulting model has similarities to the Variational Auto-Encoder presented in their paper. 
Combining RNNs with SGVB is partly inspired by the work of Justin Bayer, the first results of which were presented as workshop at NIPS 2014 \citep{bayer2014b}.

A VRAE allows to map time sequences to a latent representation and it enables efficient, large scale unsupervised variational learning on time sequences. Also, a trained VRAE gives a sensible initialisation of weights and network state for a standard RNN. In general, the network states are initialised at zero, however \citet{pascanu2012difficulty} have shown that the network state is a large factor in explaining the exploding gradients problem. Initializing a standard RNN with weights and a network state obtained from the VRAE will likely make training more efficient and will possibly avoid the exploding gradients problem and enable better scores.

\section{Methods}

\subsection{SGVB}
Stochastic Gradient Variational Bayes (SGVB) as independently developed by \citet{kingma2013auto} and \citet{rezende2014stochastic} is a way to train models where it is assumed that the data is generated using some unobserved continuous random variable $z$. In general, the marginal likelihood $\int p(z)p(x|z)dz$ is intractable for these models and sampling based methods are too computationally expensive even for small datasets. SGVB solves this by approximating the true posterior $p(z|x)$ by $q(z|x)$ and then optimizing a lower bound on the log-likelihood. Similar to the nomenclature in Kingma's paper, we call $q(z|x)$ the encoder and $p(x|z)$ the decoder.

The log-likelihood of a datapoint i can be written as a sum of the lower bound and the KL divergence term between the true posterior $p(z|x)$ and the approximation $q(z|x)$, with $\theta$ the parameters of the model:

\begin{align*}
    \log p(\mathbf{x}^{(i)}) = D_{KL}(q(\mathbf{z}|\mathbf{x}^{(i)}) || p(\mathbf{z}|\mathbf{x}^{(i)})) + \mathcal{L}(\mathbf{\theta}; \mathbf{x}^{(i)})
\end{align*}

Since the KL divergence is non-negative, $\mathcal{L}(\mathbf{\theta}; \mathbf{x}^{(i)})$ is a lower bound on the log-likelihood. This lower bound can be expressed as:

\begin{align*}
    \mathcal{L}(\theta; \mathbf{x}^{(i)}) = - D_{KL}(q(\mathbf{z}|\mathbf{x}^{(i)}) || p(\mathbf{z})) + \mathbb{E}_{q(\mathbf{z}|\mathbf{x}^{(i)})} [\log p_\theta(\mathbf{x}^{(i)}|\mathbf{z})]
\end{align*}

If we want to optimize this lower bound with gradient ascent, we need gradients with respect to all the parameters. Obtaining the gradients of the encoder is relatively straightforward, but obtaining the gradients of the decoder is not. In order to solve this \cite{kingma2013auto} introduced the "reparametrization trick" in which they reparametrize the random variable $\mathbf{z} \sim q(\mathbf{z}|\mathbf{x})$ as a deterministic variable $\mathbf{z} = g(\epsilon, x)$. In our model the latent variables are univariate Gaussians, so the reparametrization is $\mathbf{z} = \mu + \sigma\epsilon$ with $\epsilon \sim \mathcal{N}(0,1)$. 

Modelling the latent variables in this way allows the KL divergence to be integrated analytically, resulting in the following estimator:
\begin{align*}
\mathcal{L}(\theta; \mathbf{x}^{(i)}) \simeq \sum_{j=1}^{J}(1 + \log((\sigma^{(i)})^2) - (\mu_j^{(i)})^2 - (\sigma^{(i)}_j)^2) + \frac{1}{L} \sum_{l=1}^{L} \log p(x^{(i)} | z^{(i,l)})
\end{align*}

For more details refer to \citet{kingma2013auto}. They also present an elaborate derivation of this estimator in their appendix.

\subsection{Model}
The encoder contains one set of recurrent connections such that the state $h_{t+1}$ is calculated based on the previous state and on the data $x_{t+1}$ of the corresponding time step. The distribution over $Z$ is obtained from the last state of the RNN, $h_{end}$, such that:
\begin{align*}
h_{t+1} &= \tanh(W_{enc}^T h_t + W_{in}^T x_{t+1} + b_{enc}) \\
\mu_z &= W_{\mu}^T h_{end} + b_{\mu} \\
log(\sigma_{z}) &= \ W_{\sigma}^T h_{end} + b_{\sigma} 
\end{align*}

Where $h_0$ is initialised as a zero vector.
 
Using the reparametrization trick, $z$ is sampled from this encoding and the initial state of the decoding RNN is computed with one set of weights. Hereafter the state is once again updated as a traditional RNN:
\begin{align*}
h_0 &= \tanh(W_z^T z + b_z) \\
h_{t+1} &= \tanh(W_{dec}^T h_{t} + W_{x}^T x_t + b_{dec}) \\
x_t &= \mbox{sigm}(W_{out}^T h_{t} + b_{out})
\end{align*}

\section{Experiments}

\subsection{Data and preprocessing}

For our experiments we used 8 MIDI files (binary data with one dimension for each pitch) of well-known 80s and 90s video game songs\footnote{Tetris, Spongebob Theme Song, Super Mario, Mario Underworld, Mario Underwater, Mariokart 64 Choco Mountain, Pokemon Center and Pokemon Surf} sampled at 20Hz. Upon inspection, only 49 of the 88 dimensions contained a significant amount of notes, so the other dimensions were removed. The songs are divided into short parts, where each part becomes one data point. In order to have an equal number of data points from each with song, only the first 520 data points from each song were used.

\subsection{Training a model}

\begin{figure}[ht!]
\begin{subfigure}{.5\textwidth}
\centering
\includegraphics[width=1\textwidth]{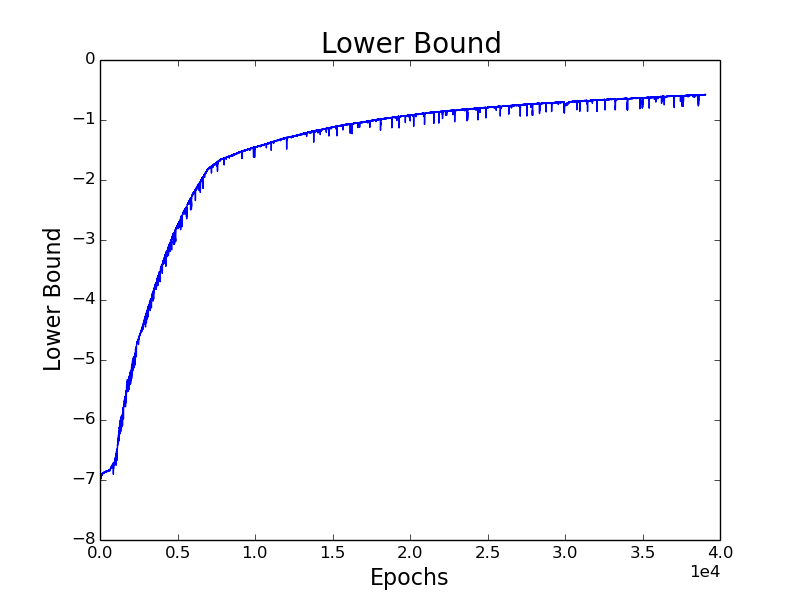}
\end{subfigure}
\begin{subfigure}{.5\textwidth}
\includegraphics[width=1.1\textwidth]{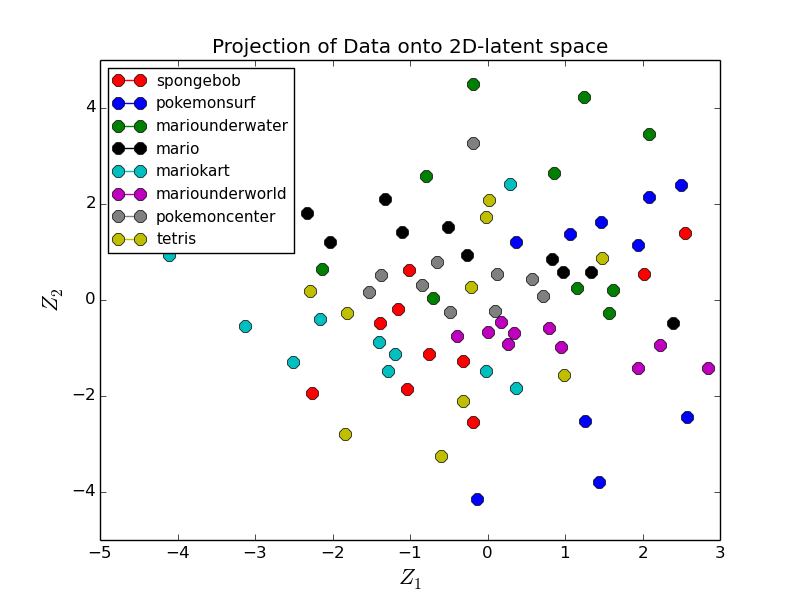}
\end{subfigure}
\caption{On the left is the lower bound of the log-likelihood per datapoint per time step during training. The first 10 epochs were cut off for scale reasons.  On the right is the organisation of all data points in latent space. Each datapoint is encoded, and visualized at the location of the resulting two-dimensional mean $\mu$ of the encoding.  "Mario Underworld" (green triangles), "Mario" (red triangles) and "Mariokart" (blue triangles) occupy the most distinct regions.}
\label{2dprojection}
\end{figure}

The choice of optimizer proved vital to make the VRAE learn a useful representation, especially adaptive gradients and momentum are important. In our experiments we used Adam, which is an optimizer inspired by RMSprop with included momentum and a correction factor for the zero bias, created by \citet{kingma2014adam}. \\
We trained a VRAE with a two-dimensional latent space and 500 hidden units on the dataset described in the last section. The songs were divided into non-overlapping sequences of 50 time steps each. Adam parameters used are $ \beta_1 = 0.05$ and $\beta_2 = 0.001$. Due to instability, the learning rate was decreased gradually during learning. The initial learning rate was to $1 \cdot 10^{-3}$ and the final learning rate was $5 \cdot 10^{-6}$. The resulting lower bound during training is shown in Figure \ref{2dprojection}.

\begin{figure}[ht!]
\begin{subfigure}{.5\textwidth}
\centering
\includegraphics[width=1\textwidth]{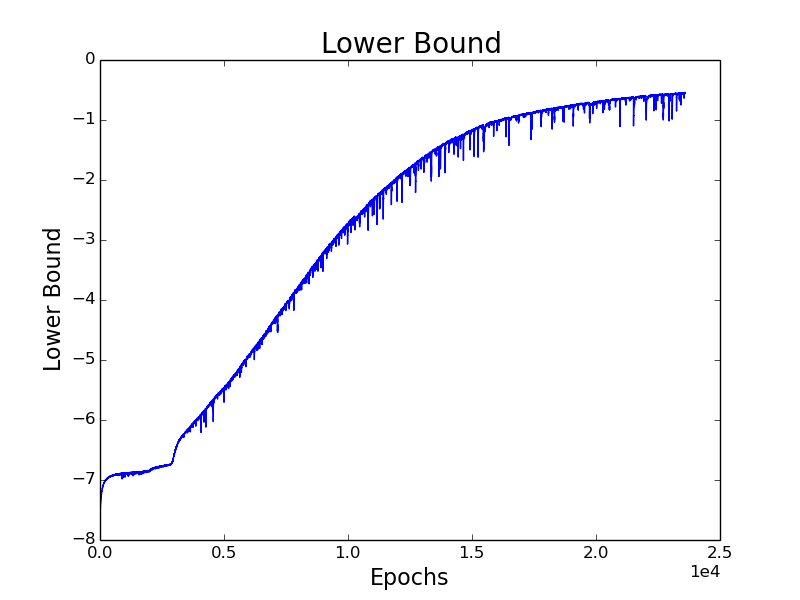}
\end{subfigure}
\begin{subfigure}{.5\textwidth}
\includegraphics[width=1.1\textwidth]{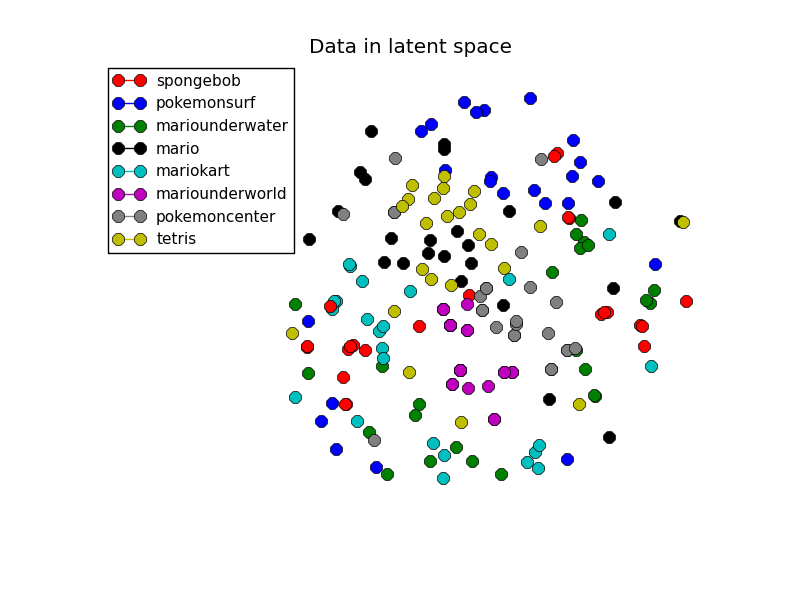}
\end{subfigure}
\caption{On the left is the lower bound of the log-likelihood per datapoint per time step during training. The first 10 epochs were cut off for scale reasons.  On the right is a visualization of the organisation of the encoded data in latent space. We calculated the 20-dimensional latent representation is calculated for each data point. The mean $\mu$ of this representation is visualized in two dimensions using t-SNE. Each color represents the data points from one song. It can be seen that for each song, the parts of that song occupy only a part of the space and the parts of some songs (e.g. "mariounderworld", in purple), are clearly grouped together. Of course, how much the parts of one song can be grouped together depends on the homogeneity of the song relative to the similarity between the different songs, as well as on how much spatial information is lost during the dimensionality reduction of t-SNE.}
\label{tsne}
\end{figure}

With a model that has only two-dimensional latent space, it is possible to show the position of each data point in latent space. The data points are only a few seconds long and can therefore not capture all the characteristics of the song. Nevertheless, Figure \ref{2dprojection} shows some clustering as certain songs occupy distinct regions in latent space. \\
A two-dimensional latent space, however, is suboptimal for modelling the underlying distribution of the data. Therefore we also trained a model with twenty latent variables. For this model, we used sequences of 40 time steps with overlap, such that the start of each data point is halfway through the previous data point. This way the model not only learns the individual data points but also the transitions between them, which enables generating music of arbitrary length. As in training the first model, Adam parameters used are $ \beta_1 = 0.05$ and $\beta_2 = 0.001$. The learning rate was $2 \cdot 10^{-5}$ and was adjusted to $1 \cdot 10^{-5}$ after $1.6 \cdot 10^4$ epochs. The resulting lower bound is shown in Figure \ref{tsne}.\\
Similar to \ref{2dprojection}, the organisation of the data in latent space using this model is shown in \ref{tsne}. In order to visualize the twenty-dimensional latent representations we used t-SNE \citep{van2008visualizing}.

\subsection{Generating data}
Given a latent space vector, the decoding part of the trained models can be used for generating data. The first model described in this chapter was trained on non-overlapping sequences of 50 time steps. Therefore, it can not be expected that generating longer sequences will yield data from the same distribution as the training data. However, since we know for each data point its latent representation in two dimensions and we can inspect their positions (see Figure \ref{2dprojection}) we use the model to interpolate between parts of different songs. The resulting music, which only lasts for a few seconds, clearly has elements of both parts. The model trained on overlapping data points was used to generate music of 1000 time steps ($\sim$50 seconds) with various (20-dimensional) latent state vectors. It is possible to obtain latent vectors by encoding a data point, or to sample randomly from latent space. Doing this creates what one might call a "medley" of the songs used for training. A generated sample from a randomly chosen point in latent space is available on YouTube \footnote{\url{http://youtu.be/cu1_uJ9qkHA}}.

\section{Discussion}
%

We have shown that it is possible to train RNNs with SGVB for effective modeling of time sequences. An important difference with earlier, similar approaches is that our model maps time sequences to one latent vector, as opposed to latent state sequences. \\
A first possible improvement over the current model is dividing each song into as many data points as possible for training (i.e. one datapoint starting at each time step) instead of data points that only have 50\% overlap. Another improvement is to reverse the order of the input, such that the the first time steps are more strongly related to the latent space than the last time steps. This will likely improve upon the length of the time dependency that can be captured, which was around 100 time steps with our current approach. Another way to train on longer time sequences is to incorporate the LSTM framework \citep{hochreiter1997long}. \\
Direct applications of our approach include recognition, denoising and feature extraction. The model can be combined with other (supervised or unsupervised) models for sequential data, for example to improve on current music genre tagging methods, e.g. \citet{sigtia2014rnn}. In addition, this method could complement current methods for supervised training of RNNs by providing initial hidden states.  

\newpage
\bibliography{iclr2015}

\begin{thebibliography}{11}
\providecommand{\natexlab}[1]{#1}
\providecommand{\url}[1]{\texttt{#1}}
\expandafter\ifx\csname urlstyle\endcsname\relax
  \providecommand{\doi}[1]{doi: #1}\else
  \providecommand{\doi}{doi: \begingroup \urlstyle{rm}\Url}\fi

\bibitem[Bayer \& Osendorfer(2014)Bayer and Osendorfer]{bayer2014b}
Bayer, Justin and Osendorfer, Christian.
\newblock Learning stochastic recurrent networks.
\newblock In \emph{NIPS 2014 Workshop on Advances in Variational Inference},
  2014.

\bibitem[Boulanger-Lewandowski et~al.(2012)Boulanger-Lewandowski, Bengio, and
  Vincent]{boulanger2012modeling}
Boulanger-Lewandowski, Nicolas, Bengio, Yoshua, and Vincent, Pascal.
\newblock Modeling temporal dependencies in high-dimensional sequences:
  Application to polyphonic music generation and transcription.
\newblock In \emph{The 29th International Conference on Machine Learning
  (ICML)}, 2012.

\bibitem[Cho et~al.(2014)Cho, van Merrienboer, Gulcehre, Bougares, Schwenk, and
  Bengio]{cho2014learning}
Cho, Kyunghyun, van Merrienboer, Bart, Gulcehre, Caglar, Bougares, Fethi,
  Schwenk, Holger, and Bengio, Yoshua.
\newblock Learning phrase representations using rnn encoder-decoder for
  statistical machine translation.
\newblock In \emph{Conference on Empirical Methods in Natural Language
  Processing (EMNLP)}, 2014.

\bibitem[Graves et~al.(2009)Graves, Liwicki, Fern{\'a}ndez, Bertolami, Bunke,
  and Schmidhuber]{graves2009novel}
Graves, Alex, Liwicki, Marcus, Fern{\'a}ndez, Santiago, Bertolami, Roman,
  Bunke, Horst, and Schmidhuber, J{\"u}rgen.
\newblock A novel connectionist system for unconstrained handwriting
  recognition.
\newblock \emph{Pattern Analysis and Machine Intelligence, IEEE Transactions
  on}, 31\penalty0 (5):\penalty0 855--868, 2009.

\bibitem[Hochreiter \& Schmidhuber(1997)Hochreiter and
  Schmidhuber]{hochreiter1997long}
Hochreiter, Sepp and Schmidhuber, J{\"u}rgen.
\newblock Long short-term memory.
\newblock \emph{Neural computation}, 9\penalty0 (8):\penalty0 1735--1780, 1997.

\bibitem[Kingma \& Ba(2014)Kingma and Ba]{kingma2014adam}
Kingma, Diederik~P and Ba, Jimmy.
\newblock Adam: A method for stochastic optimization.
\newblock \emph{ArXiv preprint arXiv:1412.6980}, 2014.

\bibitem[Kingma \& Welling(2013)Kingma and Welling]{kingma2013auto}
Kingma, Diederik~P and Welling, Max.
\newblock Auto-encoding variational bayes.
\newblock In \emph{The 2nd International Conference on Learning Representations
  (ICLR)}, 2013.

\bibitem[Pascanu et~al.(2013)Pascanu, Mikolov, and
  Bengio]{pascanu2012difficulty}
Pascanu, Razvan, Mikolov, Tomas, and Bengio, Yoshua.
\newblock On the difficulty of training recurrent neural networks.
\newblock In \emph{Proceedings of the 30th International Conference on Machine
  Learning (ICML)}, 2013.

\bibitem[Rezende et~al.(2014)Rezende, Mohamed, and
  Wierstra]{rezende2014stochastic}
Rezende, Danilo~Jimenez, Mohamed, Shakir, and Wierstra, Daan.
\newblock Stochastic backpropagation and approximate inference in deep
  generative models.
\newblock In \emph{Proceedings of the 31th International Conference on Machine
  Learning, (ICML)}, 2014.

\bibitem[Sigtia et~al.(2014)Sigtia, Benetos, Cherla, Weyde, Garcez, and
  Dixon]{sigtia2014rnn}
Sigtia, Siddharth, Benetos, Emmanouil, Cherla, Srikanth, Weyde, Tillman,
  Garcez, Artur S~d’Avila, and Dixon, Simon.
\newblock An rnn-based music language model for improving automatic music
  transcription.
\newblock In \emph{International Society for Music Information Retrieval
  Conference (ISMIR)}, 2014.

\bibitem[Van~der Maaten \& Hinton(2008)Van~der Maaten and
  Hinton]{van2008visualizing}
Van~der Maaten, Laurens and Hinton, Geoffrey.
\newblock Visualizing data using t-sne.
\newblock \emph{Journal of Machine Learning Research}, 9\penalty0
  (2579-2605):\penalty0 85, 2008.

\end{thebibliography}
\bibliographystyle{iclr2015}

\end{document}